\pdfoutput=1

\documentclass[11pt]{article}
\usepackage{graphicx}
\usepackage{acl}
\usepackage{amsmath}
\usepackage{listings}
\usepackage{enumitem}

\usepackage{times}
\usepackage{latexsym}

\usepackage{booktabs}
\usepackage{tabularx}
\usepackage{multirow}

\usepackage[T1]{fontenc}

\usepackage[utf8]{inputenc}

\usepackage{microtype}
\title{E-NER --- An Annotated Named Entity Recognition Corpus of Legal Text}

\author{Ting Wai Terence Au\textsuperscript{\textbf{1}}, Vasileios Lampos\textsuperscript{\textbf{1}} \and Ingemar J. Cox\textsuperscript{\textbf{1,2}}\\\\
    \textsuperscript{\textbf{1}} Centre for Artificial Intelligence, Department of Computer Science,\\ University College London, UK\\
    \textsuperscript{\textbf{2}} Department of Computer Science, University of Copenhagen, Denmark\\
   {\tt\small {\normalfont\{}ting.au.19{\normalfont,} v.lampos{\normalfont\}}@ucl.ac.uk{\normalfont,} ingemar@ieee.org}
}

\begin{document}

\maketitle

\begin{abstract}
Identifying named entities such as a person, location or organization, in documents can highlight key information to readers. Training Named Entity Recognition (NER) models requires an annotated data set, which can be a time-consuming labour-intensive task. Nevertheless, there are publicly available NER data sets for general English. Recently there has been interest in developing NER for legal text. However, prior work and experimental results reported here indicate that there is a significant degradation in performance when NER methods trained on a general English data set are applied to legal text. We describe a publicly available legal NER data set, called E-NER, based on legal company filings available from the US Securities and Exchange Commission's EDGAR data set. Training a number of different NER algorithms on the general English CoNLL-2003 corpus but testing on our test collection confirmed significant degradations in accuracy, as measured by the F1-score, of between 29.4\% and 60.4\%, compared to training and testing on the E-NER collection.
\end{abstract}

\section{Introduction}
\label{sec:intro}
Named Entity Recognition (NER) aims to identify names of specific objects in the world (mostly nouns with few exceptions), such as the name of a person, location and organization, which indicate possibly important phrases that readers should pay attention to. NER has been used in a variety of downstream tasks such as question answering~\cite{molla2006named}, document de-identification~\cite{stubbs2015automated,catelli2020crosslingual}, relation extraction~\cite{miwa2016end}, and machine translation~\cite{babych2003improving}. Consequently, there has been considerable work on NER using general language corpora~\cite{yadav2018survey,li2020survey} and a variety of test collections are available. Previous work has examined domain-specific NER, e.g. in finance~\cite{alvarado2015domain,alexander2021research,zhang2022finbert}, biomedical~\cite{zhou2004recognizing,wang2018comparative}, online user-generated content~\cite{tran2015semi,li2014tweet}, and legal~\cite{luz2018lener} applications, and found that the performance of domain-specific NER systems was poor if trained on general language corpora. Constructing test collections for specialist domains can be a time consuming task requiring manual annotation of a corpus. To reduce this effort there has been considerable recent interest in transfer learning, such as pre-trained language models~\cite{brown2020language,howard2018universal}. Nevertheless, there remains a need for specialist test collections whether for training or fine-tuning.

Legal text is one specialist domain where NER is of interest, due to its usefulness in assisting other legal tasks such as record linkage~\cite{dozier2010named}, court case linkage~\cite{krivz2014statistical}, contract analysis~\cite{chalkidis2017extracting}, prediction of judicial decisions~\cite{aletras2016predicting}, credit risk assessment~\cite{alvarado2015domain}, and question-answering systems~\cite{jayakumar2020rnn}. However, despite increasing interest in this sub-domain, there is no publicly available corpus for the evaluation of NER methods for legal applications. 

This paper describes E-NER, an annotated NER collection of legal documents,\footnote{E-NER data set, \href{https://github.com/terenceau2/E-NER-Dataset}{github.com/terenceau2/E-NER-Dataset}} based on publicly available legal company filings in the United States Securities and Exchange Commissions' EDGAR database. Overall, we deployed four NER models to compare classification performance when (i) trained and tested on general English, (ii) trained on general English and tested on E-NER, and (iii) trained and tested on E-NER. The results support insights from earlier work, i.e. we observed significant performance degradation when trained on general English but tested on legal text. Our experiments justify the utility of a domain-specific (legal) NER corpus.

\section{Related work}
\label{sec:prior}
The primary contribution of this paper is a legal-English test collection for NER. We do not propose a new algorithm for NER and consequently restrict our description of NER methods to those used in our experimental work.

Hidden Markov models (HMM)~\cite{rabiner1984introduction} can be used to label sequences. ~\citet{bikel1997nymble} demonstrated the application of HMM to NER. Conditional Random Fields (CRF)~\cite{lafferty2001conditional} is another sequence labelling model which improves on HMM, by relaxing the stationarity and the output independence assumptions. \citet{mccallum2003early} and \citet{sobhana2010conditional} demonstrated the application of CRF to NER.

In more recent years, pre-trained language models~\cite{qiu2020pre} and prompt-based learning~\cite{liu2021pre} have demonstrated superior performance. Bidirectional Encoder Representations from
Transformers (BERT)~\cite{kenton2019bert} is a pre-trained language model which is based on transformers~\cite{vaswani2017attention}. BERT pre-trains on a large corpus of non annotated text, performing self-supervised tasks, namely masked word prediction and next sentence pairing. BERT can facilitate transfer learning: the model parameters from the pre-training step are used during the fine-tuning step, in order for the model to learn downstream tasks such as NER~\cite{souza2019portuguese,hakala2019biomedical,li2020chinese}.

There exist publicly available annotated NER data sets for general English text, such as CoNLL-2003~\cite{sang2003introduction}, WNUT17~\cite{derczynski2017results}, and the Wikipedia gold standard corpus~\cite{balasuriya2009named}, as well as for other languages~\cite{neudecker2016open,sang2003introduction,santos2006harem,vsevvcikova2007named}. For legal domain-specific data sets, non annotated legal text is abundant, as detailed in~\citet{pontrandolfo2012legal}. For example, the pre-training of Legal-BERT~\cite{chalkidis2020legal} is performed on a corpus of non annotated documents consisting of legislation, court cases, and contracts from the UK, US, and the European Union. However, the fine-tuning of Legal-BERT is based on an annotated data set CONTRACTS-NER that is not publicly available. \citet{alvarado2015domain} annotated 8 filings from the US SEC EDGAR data set, the source of documents for our data set. The primary distinction between their work and ours is the size of the data set, 54K tokens in their data set vs. 400K tokens in ours. Furthermore, \citet{alvarado2015domain} was focused on NER in the financial (credit risk) rather than legal domain.

\citet{romanian} published a Romanian NER data set consisting of 370 legal documents, and \citet{trias2021named} created a data set consisting of header sections of court cases text (the header section will declare the parties involved in a court case). Finally, we also note that the EDGAR database has been used by ~\citet{loukas2022finer} to create an annotated data set, called FiNER, which contains over 1.1 million sentences. However, this data set is tagged with eXtensive Business Reporting Language (XBRL) tags, and it is used for numeric entity recognition.

\section{EDGAR-NER (E-NER) data set}
\label{sec:corpus}
We first describe the source documents that constitute the EDGAR-NER (E-NER) data set. We then enumerate the named entity classes, which slightly extend those used by CoNLL-2003 (CoNLL),\footnote{CoNLL-2003, \href{https://www.clips.uantwerpen.be/conll2003/ner/}{clips.uantwerpen.be/conll2003/ner/}} which is widely used in the NER community.

Financial entities, such as companies, individuals, and funds, that are registered with the United States Securities and Exchange Commission (US SEC) are required by law to submit financial filings to the Electronic Data Gathering, Analysis, and Retrieval system (EDGAR). All filings in the EDGAR data set are publicly available. There is a wide variety of different filings some of which contain almost no text, e.g. Form 3 (Initial statement of beneficial ownership of securities) or Form 4 (Statement of changes in beneficial ownership of securities). We have arbitrarily chosen the year 2010 and downloaded 52 documents.

The 52 EDGAR documents consisted of a variety of different filings. We only selected filings that contain content in the form of sentences, such as Form 10-Q, which are company quarterly reports, or Form 8-K, which are used by companies to announce major events relevant to their shareholders. The 52 documents consist of 24 different types of forms. Please see Appendix~\ref{appendix: formtype} for details. 

The filings were downloaded using the index file\footnote{This is available at \href{https://www.sec.gov/os/accessing-edgar-data}{sec.gov/os/accessing-edgar-data}.} provided by EDGAR, in the form of HTML text. Each document was pre-processed using the Python package ``Beautiful Soup'' to extract sentences. We remove: 
\begin{itemize}[wide=0pt]
    \item the SEC filing header, where the filer fills in the information in a designated space. This is indicated by the HTML tag \texttt{<SEC-HEADER>}.
    \item graphical elements, such as company logos or scanned photos. This is indicated by \texttt{<TYPE>GRAPHIC}.
    \item tables with no sentences in them. Tables are indicated by the HTML tag \texttt{TABLE}.
    \item page titles and page numbers.
    \item figures and plots.
    \item any XBRL (eXtensible Business Reporting Language) instance.
\end{itemize} 
An illustration of what elements we removed or kept in an example filing is shown in Appendix~\ref{appendix: filingexample}. After preprocessing the 52 documents, we split the document into sentences by identifying the line breaks in the document, and using the sentence tokenizer from the Python NLTK package. In total, we identified 11{,}696 sentences that required annotation.

Annotation of the collection was performed by the first author. Note that we did attempt to outsource the annotation to a commercial crowdsourcing platform. We provided instructions, including the definitions of the named entity classes and the tagging guidelines. Each document was assigned to 3 crowd workers to independently label so as to ensure the correctness of the labels. However, we found that there were significant discrepancies in the labels provided. While we acknowledge that this variation may have been due to our instructions being poor, it is our opinion that the task has a significant difficulty for a non-expert.

\begin{table}[t]
\setlength{\belowrulesep}{0pt}
\setlength{\aboverulesep}{0pt}
\renewcommand{\arraystretch}{1.2}
\centering
\begin{tabular}{cc} 
 \toprule
 \textbf{CoNLL}  & \textbf{E-NER}\\
 \toprule
 Location & Location\\
  \hline
 Person & Person\\
 \hline
 Organization & Business \\
              & Goverment \\
              & Court\\
 \hline
 Miscellaneous & Legislation/Act\\
               & Miscellaneous\\
 \bottomrule
\end{tabular}
 \caption{Named entities used in the CoNLL and E-NER data sets and their pairing in the two classficiation frameworks}
 \label{tab:neclass}
\end{table}

The CoNLL-2003 data set defines 4 classes of named entities (and the class ``Outside'' for non-named entities)\footnote{See \href{https://www.cnts.ua.ac.be/conll2003/ner/annotation.txt}{cnts.ua.ac.be/conll2003/ner/annotation.txt}} as enumerated in Table~\ref{tab:neclass}. For our data set we annotated the filings with 7 named entity classes as shown in Table~\ref{tab:neclass}. 
We note that there is no consensus on the appropriate labeling of named entities for the legal domain, with various authors~\cite{dozier2010named,cardellino,leitner2019fine} proposing related but different classifications. Our class labels were chosen in consultation with a legal company (Clifford Chance LLP). Note, however, that for the experimental results reported in Section~\ref{sec:experiments}, we used the same categories as CoNLL-2003, merging and matching categories as shown in Table~\ref{tab:neclass}. E-NER follows the same file format conventions as CoNLL.

\begin{table*}[t]
\setlength{\belowrulesep}{0pt}
\setlength{\aboverulesep}{0pt}
\renewcommand{\arraystretch}{1.2}
\centering
\begin{tabular}{lccccc} 
 \toprule
 \textbf{Data set} & \textbf{Tokens} & \textbf{Sentences} & \textbf{Avg. words / sentence}& \textbf{NE phrases} & \textbf{Avg. tokens / NE}\\
 \toprule
 CoNLL train & 204{,}563    & 14{,}986 & 13.7   & 23{,}498  & 1.45 \\
  \hline
  CoNLL val. & 51{,}578     & 3{,}466  & 14.9   & 5{,}942   & 1.45\\
   \hline
 CoNLL test  & 46{,}666     & 3{,}684  & 12.7   & 5{,}648   & 1.44\\
  \hline
 E-NER      & 403{,}673    & 11{,}696 & 34.5   & 8{,}821   & 2.68\\
 \bottomrule
\end{tabular}
\caption{Basic statistics of the CoNLL and E-NER data sets}
\label{tab:dataset} 
\end{table*}

Table~\ref{tab:dataset} provides a statistical comparison between the E-NER and CoNLL-2003 data sets. We see that while the number of tokens in the E-NER data set exceeds that of CoNLL (by combining the training, validation, and test sets), the number of NE phrases is considerably smaller (8{,}821 for E-NER, compared to 35{,}088 CoNLL combined). We also observe that the CoNLL data set has considerably more sentences (22{,}136 vs. 11{,}696) and that these sentences are much shorter (13.7 words vs. 34.5 words per sentence). The number of tokens constituting a NE is also shorter in CoNLL (1.45 vs. 2.68).

\begin{table}[b!]
\centering
\setlength{\belowrulesep}{0pt}
\setlength{\aboverulesep}{0pt}
\renewcommand{\arraystretch}{1.2}
\begin{tabular}{lccc}
    \toprule
    \textbf{Model}  & \textbf{G to G} & \textbf{G to L} & \textbf{L to L}\\
    \toprule
    Baseline        & .596 & .136 & .491\\
    \hline
    HMM             & .622 & .148 & .401\\
    \hline
    CRF             & .820 & .216 & .902\\
    \hline
    BERT            & .905 & .611 & .961\\
    \bottomrule
\end{tabular}
\caption{F1-scores of different models when trained (or fine-tuned) and tested on different data sets. In the column headers, the first entry is the training data set (or data set to fine-tune on), and the second is the test data set. \textbf{G} denotes a general data set for NER (here CoNLL), and \textbf{L} denotes a legal data set (here E-NER). For the column \textbf{L to L}, we perform 5-fold cross-validation.}
\label{results}
\end{table}

\section{Experiments}
\label{sec:experiments}

To verify the need for a legal NER collection, we evaluated the performance of four NER methods by (i) training and testing on a general English collection (CoNLL), (ii) training on general English, but testing on our legal collection (E-NER), and (iii) training and testing on our E-NER collection.

The CoNLL collection is subdivided into train, validation, and test partitions, as indicated in Table~\ref{tab:dataset}. When training and testing using E-NER, we performed $k$-fold cross-validation. Since the size of the train and test data sets in CoNLL-2003 has a ratio of approximately 4:1, we chose $k=5$. We report the micro-F1 score.

\subsection{CoNLL-2003 workshop baseline model}
The baseline model records all the NE phrases in the training set. During testing, phrases are matched against these learned NE phrases and labeled accordingly (i.e. there is no generalisation). If a phrase in the dictionary has multiple NE classes associated to it, the one with the highest frequency is used.

\subsection{Hidden Markov Model}
Our HMM implementation follows the same approach as proposed by ~\citet{morwal2012named}. The NE tags are treated as the hidden states, and the tokens are treated as the observations.

\subsection{Conditional Random Fields}
Our CRF implementation is similar to the one proposed by ~\citet{mccallum2003early}. However, we did not use lexicons or other reference corpora to assist our CRF models to identify names of countries, companies, and surnames. Our choice of feature functions is hand-crafted, and consists of (i) the current word, (ii) the first and last 2 letters of the current word, (iii) the capitalization of the word, and (iv) the above 3 features for the word to the left and to the right of the current word.

\subsection{BERT}
We used a pre-trained version of BERT.\footnote{BERT, \href{https://huggingface.co/bert-base-uncased}{huggingface.co/bert-base-uncased}} In our experiments, we fine-tuned BERT using Hugging Face's transformer package.\footnote{Available at \href{https://github.com/huggingface/transformers/tree/main/examples/pytorch/token-classification}{github.com/huggingface/transformers/{\newline}tree/main/examples/pytorch/token-classification}.}

\subsection{Results}
In Table~\ref{results}, we present the F1-score for the aforementioned NER models when we train and test them on different data sets. In the columns, the first entry in the brackets shows the data set used for training (or fine-tuning), and the second entry shows the test data sets. 

When we train and test the models on the CoNLL corpus, F1-scores range from 59.6\% to 90.5\%. However, when we train on CoNLL and test on E-NER, F1-scores degrade significantly, ranging from 13.6\% to 61.1\%. When training and testing using the E-NER collection the F1-scores range from 49.1\% to 96.1\% which consistutes a significant improvement over training using the CoNLL data set. Interestingly, we observe that the dictionary baseline and HMM models perform similarly or worse on legal text compared to their performance on general English. Conversely, for the more advanced CRF and BERT models, performance on legal text exceeds that for general English. It is unclear whether this is principally due to differences in the models, or differences in the test collections. Nevertheless, experimental results support earlier work indicating degradation in performance when NER methods are trained on general English but applied to the legal domain.  

\section{Conclusions and future work}
\label{sec:conclusion}
This paper describes the publicly available E-NER data set, derived from  company filings from the US SEC EDGAR data set. The collection contains over 400,000 tokens, and as such, is of similar size to the CoNLL-2003 collection. However, the number of NE phrases (almost 9,000) is only about 25\% of the number of NE phrases in the CoNLL corpus. In part, this reflects the statistical differences between general and legal English, where we observed that the sentence length for legal English (34.5 words) is much larger than for general English (13.7), and that the token length of a NE in legal text is longer (2.68 tokens compared to 1.45). In addition, the fact that E-NER encompasses only 52 documents from EDGAR might also contribute to this discrepancy.

Our experimental results compared the performance of four NER methods when trained and tested on combinations of general and legal English. Our results reaffirm that for legal NER in-domain performance is significantly degraded when training without using specific in-domain data.

There is a number of potential future research directions. First, there is a variety of legal specialities, e.g. finance, civil litigation, and criminal law. Further work is needed to investigate how NER models perform in various legal sub-domains -- how diverge and large should annotated corpora be for legal NER? To this end, we plan to create annotated datasets for other types of legal documents, such as court proceedings or contracts. In addition, the evaluation of NER models using state-of-the-art methods and language models in legal NLP might unveil more informative results and drive potential methodological improvements.

\section*{Acknowledgements}
T.W.T.A. and I.J.C. would like to thank Clifford Chance LLP for the financial support and for providing guidance with respect to requirements from the legal community.

\bibliography{main}

\clearpage

\appendix

\section{Tables}
\label{appendix: formtype}

\begin{table}[h!]
\centering
\setlength{\belowrulesep}{0pt}
\setlength{\aboverulesep}{0pt}
\renewcommand{\arraystretch}{1.2}
\begin{tabular}{lccc} 
 \toprule
 \textbf{Form types} &  \textbf{Count} &   \textbf{Form types} &  \textbf{Count} \\
 \toprule
  497K & 6&   DEFA14A & 1\\
  \hline
  8-K & 6&  N-CSR & 1\\
  \hline
  10-Q & 5 &   POSASR & 1\\
  \hline
  425 & 4 &   PRE 14C & 1\\
  \hline
  N-Q & 3 &   SC 13D & 1\\
  \hline
  11-K & 3 &   SC 13DA & 1\\
  \hline
  424B3 & 3 &   S-3 & 1\\
  \hline
  CORRESP & 2 &   S-4 & 1\\
  \hline
  DEF 14A & 2 &   S-8 & 1\\
  \hline
  10-K & 2 &   10-KA & 1\\
  \hline
  40-17G & 2 &   424B5 & 1\\
  \hline
  497 & 2 &   40-APPA & 1\\
  \bottomrule
\end{tabular}
\caption{Type of forms in the E-NER data set}
\label{tab:formtypes} 
\end{table}

\section{Example filing}
\label{appendix: filingexample}

\begin{figure*}
    \centering
    \includegraphics[scale=0.36]{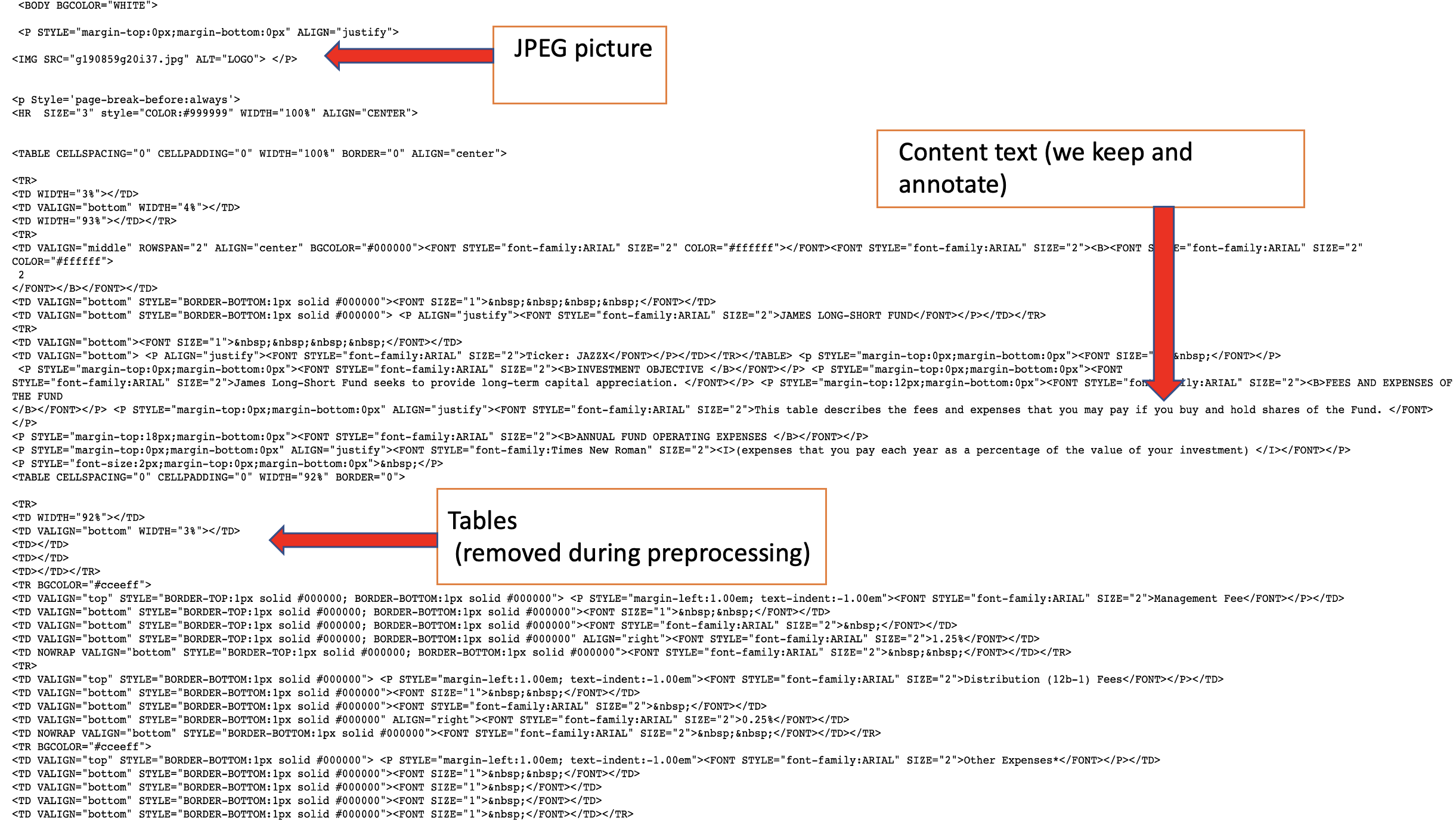}
    \caption{Raw HTML of an example filing, downloaded from the EDGAR database.}
    \label{fig:rawtext}
\end{figure*}

\begin{figure*}
    \centering
    \includegraphics[scale=0.36]{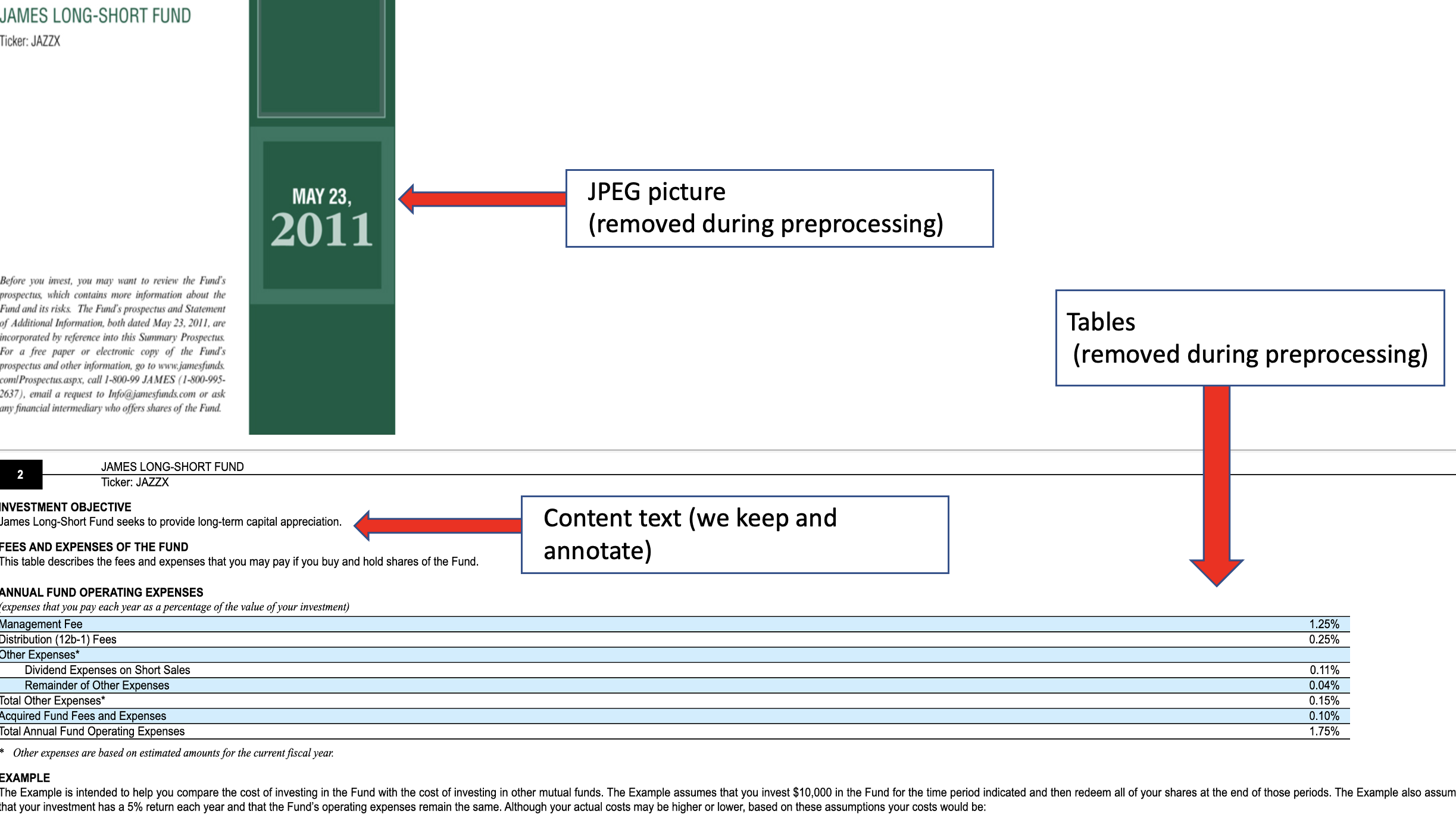}
    \caption{The rendered version of the filing.}
    \label{fig:htm}
\end{figure*}

\begin{figure*}
    \centering
    \includegraphics[scale=0.3]{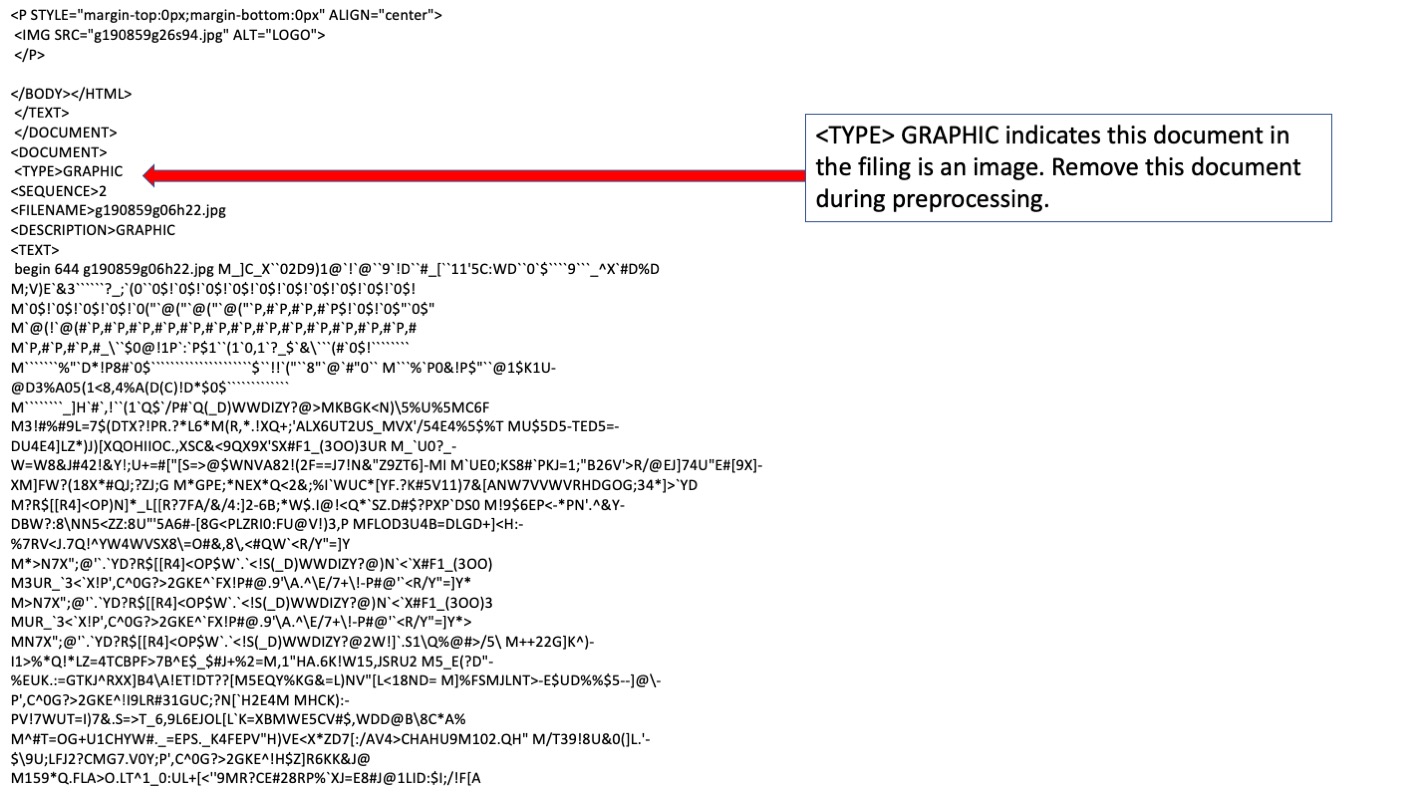}
    \caption{Raw HTML of an example filing, where one of the documents uploaded is an image.}
    \label{fig:rawtext2}
\end{figure*}

An example filing in the E-NER data set, in the form of the HTML and its rendered version, is shown in Figure ~\ref{fig:rawtext} and ~\ref{fig:htm}. Figure ~\ref{fig:rawtext2} shows an image element in this filing, which we remove during preprocessing. This filing's CIK number is 0001045487. The accession number is 000119312511147903. The URL to this filing is 
\href{https://www.sec.gov/Archives/edgar/data/1045487/000119312511147903}{sec.gov/Archives/edgar/data/1045487/0001193125{\newline}11147903}.

\end{document}